\documentclass{article}
\usepackage{spconf,epsfig}

\usepackage{amsmath,bm}
\usepackage{amssymb}
\usepackage[numbers]{natbib}

\usepackage[pagebackref=true,breaklinks=true,letterpaper=true,colorlinks,bookmarks=false]{hyperref}
\begin{document}\sloppy

\def\x{{\mathbf x}}
\def\L{{\cal L}}

\title{Classifying Online Dating Profiles on Tinder using FaceNet Facial Embeddings}
%
\name{Charles F. Jekel (cjekel@ufl.edu; cj@jekel.me) and Raphael T. Haftka}
\address{Department of Mechanical \& Aerospace Engineering - 
University of Florida - Gainesville, FL 32611}

\maketitle

\begin{abstract}
A method to produce personalized classification models to automatically review online dating profiles on Tinder, based on the user's historical preference, is proposed. The method takes advantage of a FaceNet facial classification model to extract features which may be related to facial attractiveness. The embeddings from a FaceNet model were used as the features to describe an individual's face. A user reviewed 8,545 online dating profiles. For each reviewed online dating profile, a feature set was constructed from the profile images which contained just one face. Two approaches are presented to go from the set of features for each face to a set of profile features. A simple logistic regression trained on the embeddings from just 20 profiles could obtain a 65\% validation accuracy. A point of diminishing marginal returns was identified to occur around 80 profiles, at which the model accuracy of 73\% would only improve marginally after reviewing a significant number of additional profiles. 
\end{abstract}
\section{Introduction}
Online dating has become a commonplace in today's society. The top grossing iOS application in September 2017 was an online dating service named Tinder. Users of online dating services are generally expected to spend a significant amount of time filtering through the profiles of potential partners. This study investigates whether a pattern in facial features can be used to filter online dating profiles on Tinder.


Tinder was selected for this study because of the popularity of the application \cite{Tyson2016}. The mobile dating application allows users to browse dating profiles of nearby singles. Users are presented with a single profile at a time. At first glance, the user can see the profile picture, first name, and age of an individual on Tinder. A user must then decide whether they \textit{like} or \textit{dislike} the profile on the spot. The user can view additional information such as an optional biography or extra pictures, but no new profiles can be reviewed until a decision is made. When two individuals have liked each other, they are presented with a notification and the opportunity to message each other. 

A custom application was developed to interface with Tinder. The intention of the application was to allow a user to like and dislike profiles while recording everything about the profiles. The profile information contains the mandatory requirements from tinder, which include the profile's name, age, and pictures. Every Tinder profile includes at least one image. Additionally, it is optional to include a biography, current job, and current school. All of this information is stored along with the like or dislike verdict in a database. 

A method for obtaining personalized classification models from a user's historical preferences is presented. The method takes advantage of recent advancements in computer vision related to facial detection, facial classification, and facial attractiveness. The results of a personalized classification model for a user's reviewed Tinder profiles are then presented.



\section{Related Works}
Related work focuses on literature regarding modeling facial attractiveness. Attractiveness is known to play a key role in online dating profiles \cite{Tyson2016}, \cite{Abramova2016}. 

There has been substantial work in literature to predict facial attractiveness with computer models. Traditionally, facial attractiveness has been modeled with some sort of eigenface transformation. \citet{Eisenthal2006} were able to find a correlation of facial attractiveness using an eigenface transformation on a small data set of 92 images. These eigenface transformations are run through a principle component analysis to determine the important facial features \cite{MU201359}, \cite{Chua2013}.  

Convolutional neural networks (CNN) have become popular for image processing in recent years. \citet{Gray2010} showed that a CNN trained to predict facial attractiveness could extract more meaningful features than an eigenface transformation. Consequently, facial features related to attractiveness could be extracted automatically without identifying landmarks using the CNN.

The recent works of \citet{Rothe2016} demonstrated that a CNN can be trained on a large data set to predict facial attractiveness. The training set contained the faces of women who were liked or disliked by male participants. Personalized predictions were made based on the historical preferences of 8,560 heterosexual male users. The model only considered the first profile image from each profile. The results of the personalized predictions were impressive, with a mean accuracy of 75\% for the male users. 

\citet{Zang2017} used a pre-trained VGG-Face CNN descriptor to predict the facial attractiveness of male and female users for an online dating site. They demonstrated that features from a facial classification model could be used to predict facial attractiveness. The model in \citet{Rothe2016} was trained on a data set related to facial attractiveness, while the model used in \citet{Zang2017} was trained on faces for the purpose of classification. These studies highlight that a large scale facial classification model is useful to predict facial attractiveness. The VGG-Face CNN used was created by \citet{Parkhi15} and scores an impressive 98.95\% accuracy on the Labeled Faces in the Wild (LFW) database \cite{LFWTech}.

%

These works focused solely on rating individual photos, but have not progressed to a usable model that likes or dislikes complete online dating profiles. The work presented in this paper strives to close this gap.

\section{Our Methodology}

The methodology proposed here attempts to classify an online dating profile as either a like or dislike. Two different approaches are proposed to consolidate multiple facial features from the images in a profile into a single vector of features that describes the profile. Like the related works of \cite{Rothe2016}, \cite{Zang2017}, the last layer of a CNN was used as the facial features for each face. A new implementation of the FaceNet classification model first described by \citet{Schroff_2015_CVPR} is used with a slightly higher LFW score than used in \citet{Zang2017}. 

The detection of profile images that contain only one face per image was automated using computer vision techniques. These faces were fed into a FaceNet model to extract the facial features as embeddings. A set of embeddings for reviewed online dating profiles was used to train a personalized classification model. 

The major assumptions of the purposed method are as follows: 1) An online dating profile can be reviewed using only the profile images; 2) The face of the individual profile can be found from the profile pictures that contain  only one face per image; 3) Images with more than one face in an online dating profile can be ignored. Profiles that can't be identified to a single face can be rejected; 4) A pattern exists in the faces of individuals who were liked or disliked by a user reviewing online dating profiles; 5) A trained FaceNet model can be evaluated on new faces to extract the facial features of the individuals.

\subsection{FaceNet implementation}
A Python library called facenet was used to calculate the facial embeddings of the dating profile pictures. These embeddings are from the last layer of a CNN, and can be thought of as the unique features that describe an individual's face. The facenet library was created by Sandberg as a TensorFlow implementation of the FaceNet paper by \citet{Schroff_2015_CVPR}, with inspirations from \cite{Parkhi15,Wen2016,amos2016openface}. The library uses the MIT license and is available online at \url{https://github.com/davidsandberg/facenet}. 

There are pre-trained facenet models available online. The models have been validated on the LFW database \cite{LFWTech}. The current best model has a LFW accuracy of 99.2\% and was trained as a classification model on a subset of the MS-Celeb-1M database \cite{guo2016}. The model's architecture follows the Inception-ResNet-v1 network as described by \citet{Szegedy2016}. The facenet library includes an implementation of Multitask CNN (MTCNN) by \citet{Zhang2016} to detect facial landmarks, which was used to create training faces as 182x182 pixel images from the MS-Celeb-1M database.

The facenet model turns a color image of a face into a vector of 128 floating point numbers. These 128 embeddings can be used as features for classification or clustering \cite{Schroff_2015_CVPR}. The facenet library includes a script to calculate the embeddings from images of faces using a pre-trained model.


\subsection{Classification methodology}
Classification models were determined for two different approaches on the embeddings. One approach considered all of the embeddings, from the images containing just one face, in the dating profile. These embeddings were used to describe the entire profile. The other approach rather considered the average embedding values across the images. Again, only images containing exactly one face were considered.

The first approach used the 128 embeddings from each image as the features of the profile. The embeddings from the images of the profiles can be described as the vectors of
\begin{align}
\bm{i}_1 &= [x_1, x_2, \cdots, x_{128}] \\
\bm{i}_2 &= [y_1, y_2, \cdots, y_{128}] \\
\vdots &= \vdots \\
\bm{i}_n &= [z_1, z_2, \cdots, z_{128}] 
\end{align}
for $n$ number of profile images. Then a single vector of embeddings can be constructed for the profile as 
\begin{align}
\bm{i}_p &= [\bm{i}_1, \bm{i}_2, \cdots, \bm{i}_n]
\end{align}
where $\bm{i}_p$ is a vector containing $128n$ values. 

The second approach considered the average embedding value of the facial images. Thus a profile with one facial image would have 128 unique embeddings. A profile could be described as
\begin{align}
\bm{i}_1 &= [x_1, x_2, \cdots, x_{128}] \\
\bm{i}_2 &= [y_1, y_2, \cdots, y_{128}] \\
\vdots &= \vdots \\
\bm{i}_f &= [z_1, z_2, \cdots, z_{128}] 
\end{align}
where $\bm{i}_f$ is the vector of embedding from the $f$ image in the profile. Then an average embeddings could be calculated as 
\begin{align}
\bm{i}_{\text{avg}} &= \begin{bmatrix}
\frac{x_1 + y_1 +\cdots + z_1}{f} \\
\frac{x_2 + y_2 +\cdots + z_2}{f} \\
\vdots \\
\frac{x_{128} + y_{128} +\cdots + z_{128}}{f} \\
\end{bmatrix}
\end{align}
where $\bm{i}_{\text{avg}}$ is a vector with the same size as the number of embeddings calculated. 

Calculating the facial embeddings from a user's reviewed online dating profiles is computationally inexpensive, as the calculation is simply a function evaluation on a pre-trained CNN. Then, classification models were trained using either $\bm{i}_p$ or $\bm{i}_{\text{avg}}$ as the input features. Personalized classification models could be constructed based on the preference from an individual's historically reviewed online dating profiles.

\section{Experimental Results}
A heterosexual male used the custom application with the intention of finding a romantic partner. The reviewing of tinder profiles went on for a month, but stopped early because the user found a girlfriend in the process. It may be important to mention that males may have different online dating tendencies than females \cite{Tyson2016, Abramova2016}. The user took about one hour to review 100 profiles. In the end, a data set was created which reviewed 8,545 tinder profiles. The user liked a total of 2,411 profiles. Additionally, the data set contains 38,218 images from the profiles browsed. Each image has a resolution of 640x640 pixels (px). 

The results were split into two categories. The first subsection presents the results of the data set after pre-processing was performed. The data set was transitioned from complete online dating profiles to a data set of faces for each profile. The faces were then run through a FaceNet model to extract the embeddings for each face. The second section then presents the results of classifying these embeddings for the two proposed input dimensions.

\subsection{Data set after pre-processing}
The MTCNN described by \cite{Zhang2016} was used to detect and box faces from the 640x640~px profile images. Faces were selected with a minimum size of 60x60~px and a threshold of $0.8$. Profile images that contained just one face were extracted and re-sized. A profile that did not contain a single image with only one face, was immediately removed. There were 24,486 images that contained only one face in the image (according to the MTCNN). Fortunately 8,130 profiles of the 8,545 reviewed (or 95.1\%) contained at least one uniquely identifiable face. The images containing just one face were cropped to 182x182~px images with a margin of 44~px around the face. A face at the minimum size was enlarged, while larger faces were reduced in size.  

\begin{figure}[hbt!]
\centering
\includegraphics[width=0.09\textwidth]{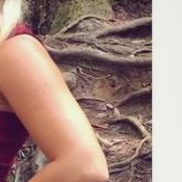}
\includegraphics[width=0.09\textwidth]{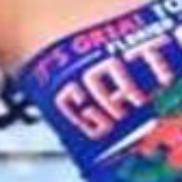}
\includegraphics[width=0.09\textwidth]{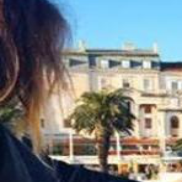}

\includegraphics[width=0.09\textwidth]{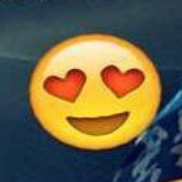}
\includegraphics[width=0.09\textwidth]{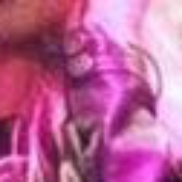}
\includegraphics[width=0.09\textwidth]{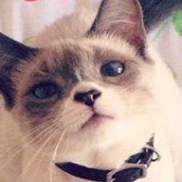}
\caption{Examples of false positives which the MTCNN identified as human faces.}
\label{fig:fp}
\end{figure}

The MTCNN results were impressive, despite the substantial amount of noise in the images. Noise includes everything from sunglasses, hats, and scarfs to Snapchat filters. For example, a particular popular Snapchat filter applies the ears, nose, and mouth of a dog to an individual's face. The MTCNN appeared to work well despite the noise in the data. There was a limited number of false positives, of which a few are presented in Fig.~\ref{fig:fp}. The false positives were not removed from the training set, as the noise they provide may be useful to construct a robust classifier. The true rate of false positives and false negatives was not studied, as the locations of faces in the original 38,218 images were not recorded. 

The embeddings were calculated for the faces from the 8,130 profiles using the FaceNet implementation described.  The average profile reviewed had 3.01 images of a single face, with a standard deviation of 1.34. Ten was the maximum number of images for a profile in the new data set. Thus $\bm{i}_p$ was a vector of $128\times 10$ in length. Profiles with fewer than ten images would have zeros in place of the missing images. Essentially a profile with just one facial image would have 128 unique embeddings and 1,152 zeros, a profile with two facial images would have 256 unique embeddings and 1,024 zeros, and so forth. The other input feature $\bm{i}_\text{avg}$ was calculated for each profile. The supplementary material includes the two input dimensions ($\bm{i}_p$ and $\bm{i}_\text{avg}$) with binary labels to show whether the profile was either liked or disliked. 

\subsection{Classification models}

In order to build a reasonable classification model, it was important to demonstrate how many profiles were required to be reviewed. Classification models were trained using various fractions of the entire data, ranging from 0.125\% to 95\% of the 8,130 profiles. At the low end, just 10 profiles were used to train the classification model, while the remaining 8,120 profiles were used to validate the trained classification model. On the other spectrum, classification models were trained using 7,723 profiles and validated on 407 profiles. 


The classification models were scored on accuracy, specifically the number of correctly classified labels over the number of profiles. The training accuracy refers to the accuracy in the training set, while the validation accuracy refers to the accuracy in the test set. 

This data set suffers from a class imbalance, as only 28\% of the total Tinder profiles reviewed were liked. The classification models were trained assuming a balanced class. A balanced class indicates that each profile considered had the same weight, regardless of whether the profile was liked or disliked. The class weight can be user dependent, as some users would value correctly liking profiles more than incorrectly disliking profiles. 

A like accuracy was introduced to represent the number of correctly labeled liked profiles out of the total number of liked profiles in the test set. Complementary, a dislike accuracy was used to measure the disliked profiles predicted correctly out of the total number of disliked profiles in the test set. A model that disliked every single profile, would have a 72\% validation accuracy, a 100\% dislike accuracy, but a 0\% like accuracy. The like accuracy is the true positive rate (or recall), while the dislike accuracy is the true negative rate (or specificity).

Various classification models were fit using either $\bm{i}_p$ or $\bm{i}_\text{avg}$ as the input. Scikit-learn was used to fit logistic regression and support vector machines (SVM) \cite{scikit-learn}, while Keras and TensorFlow were used to fit various neural networks \cite{chollet2015keras,tensorflow2015-whitepaper} to the embeddings. 

\begin{figure}[hbt!]
\centering
\includegraphics[width=0.45\textwidth]{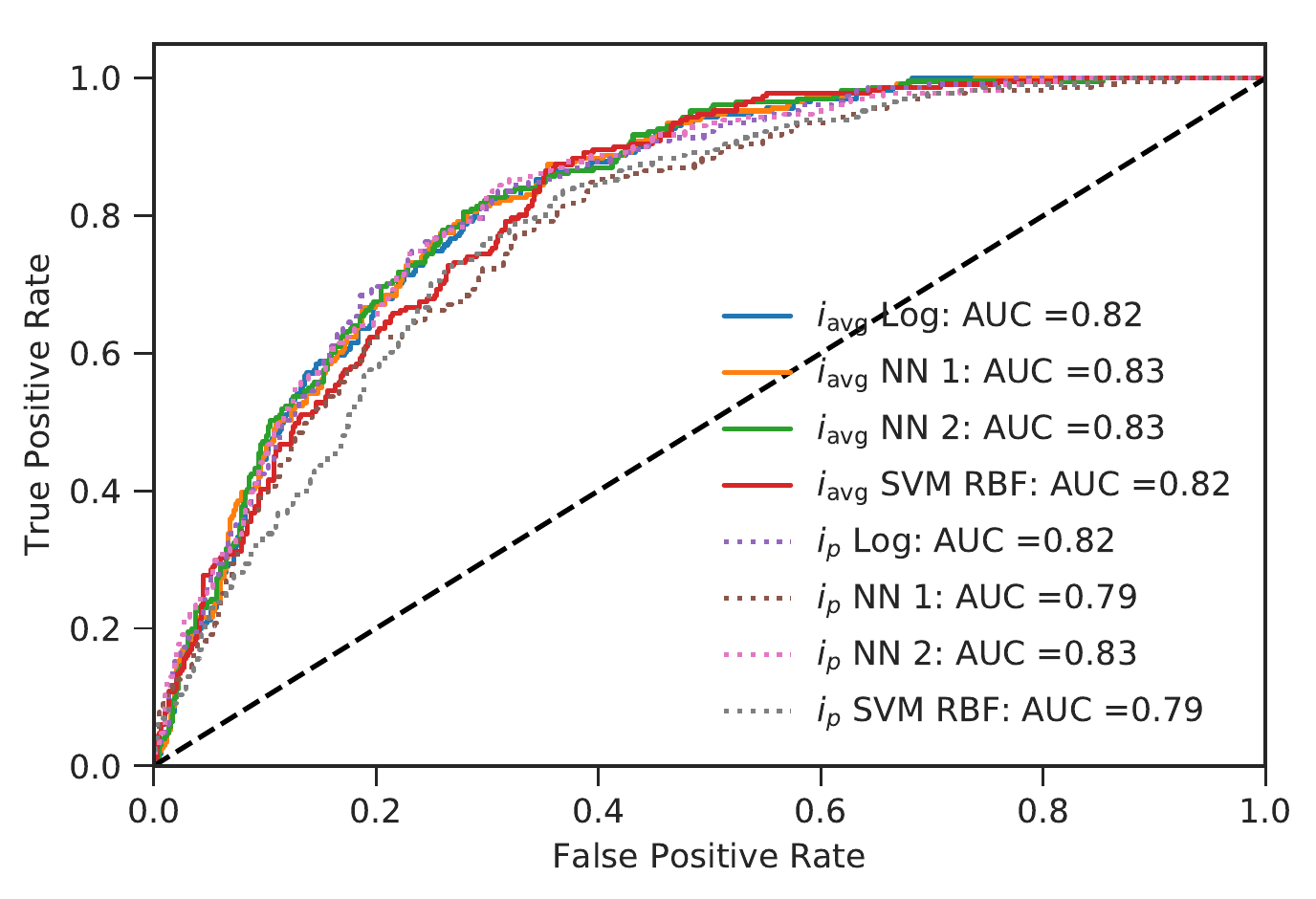}
\caption{Receiver operating characteristic (ROC) and area under curve (AUC) for various classification models using a 10:1 train:test split.}
\label{fig:roc}
\end{figure}

The receiver operating characteristic (ROC) for logistic regression (Log), neural network (NN), and SVM using radial basis function (RBF) are presented in Fig.~\ref{fig:roc}. Two different layer configurations of neural networks are presented for each input dimension as NN~1 and NN~2. Additionally, the area under curve (AUC) for each classification model is presented. The complete input dimension feature of $\bm{i}_p$ did not appear to offer any advantages over $\bm{i}_\text{avg}$ when considering AUC. A neural network had the best AUC score of 0.83, but it was only slightly better than a logistic regression with an AUC score of 0.82. This ROC study was performed using a random 10:1 train:test split (training on 7,317 and validation on 813 profiles).

\begin{figure}[hbt!]
\centering
\includegraphics[width=0.40\textwidth]{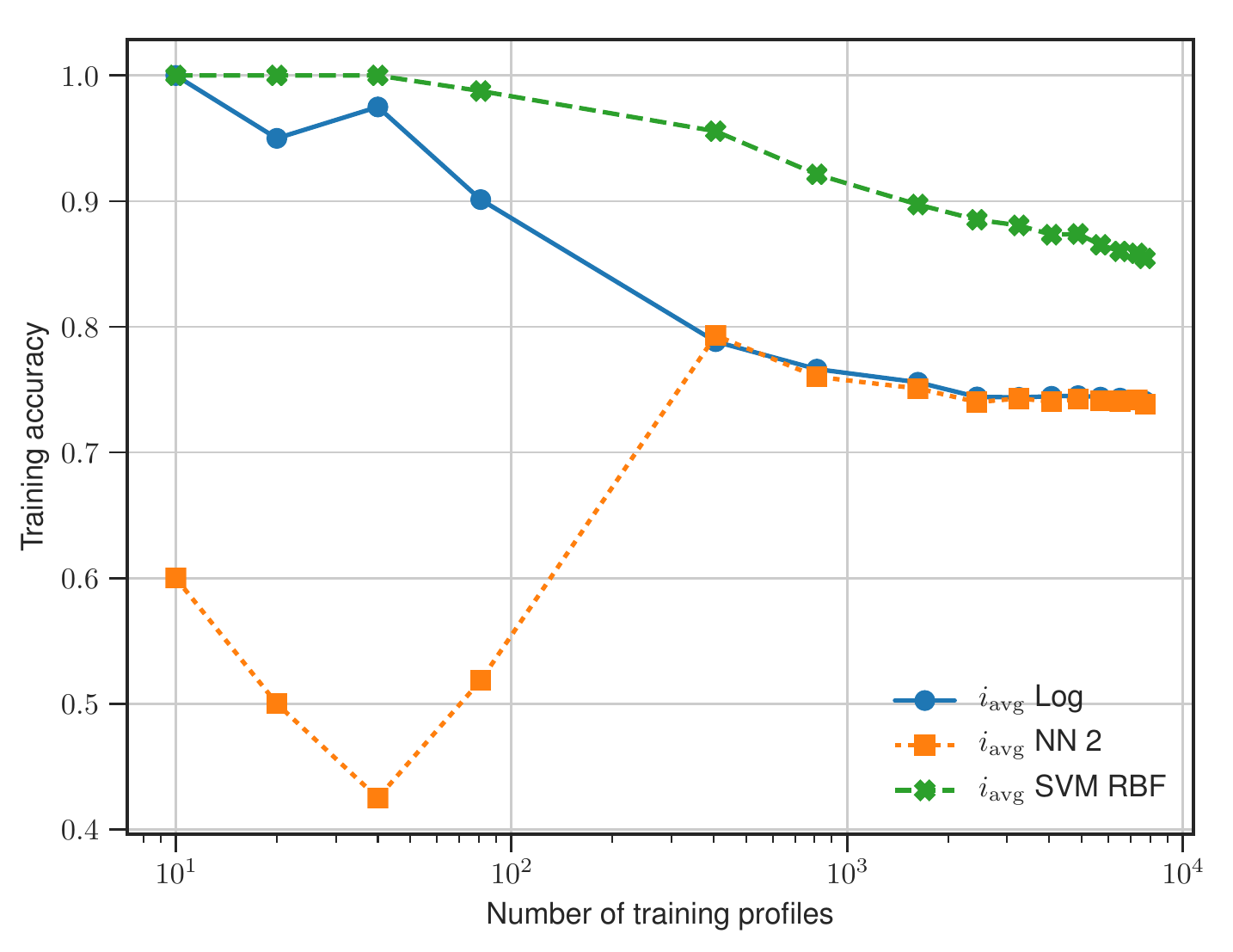}
\caption{Accuracy of the classification models evaluated on the training data (number of correctly classified training profiles / total number of training profiles) vs the number of training profiles.}
\label{fig:train}
\end{figure}

\begin{figure}[hbt!]
\centering
\includegraphics[width=0.40\textwidth]{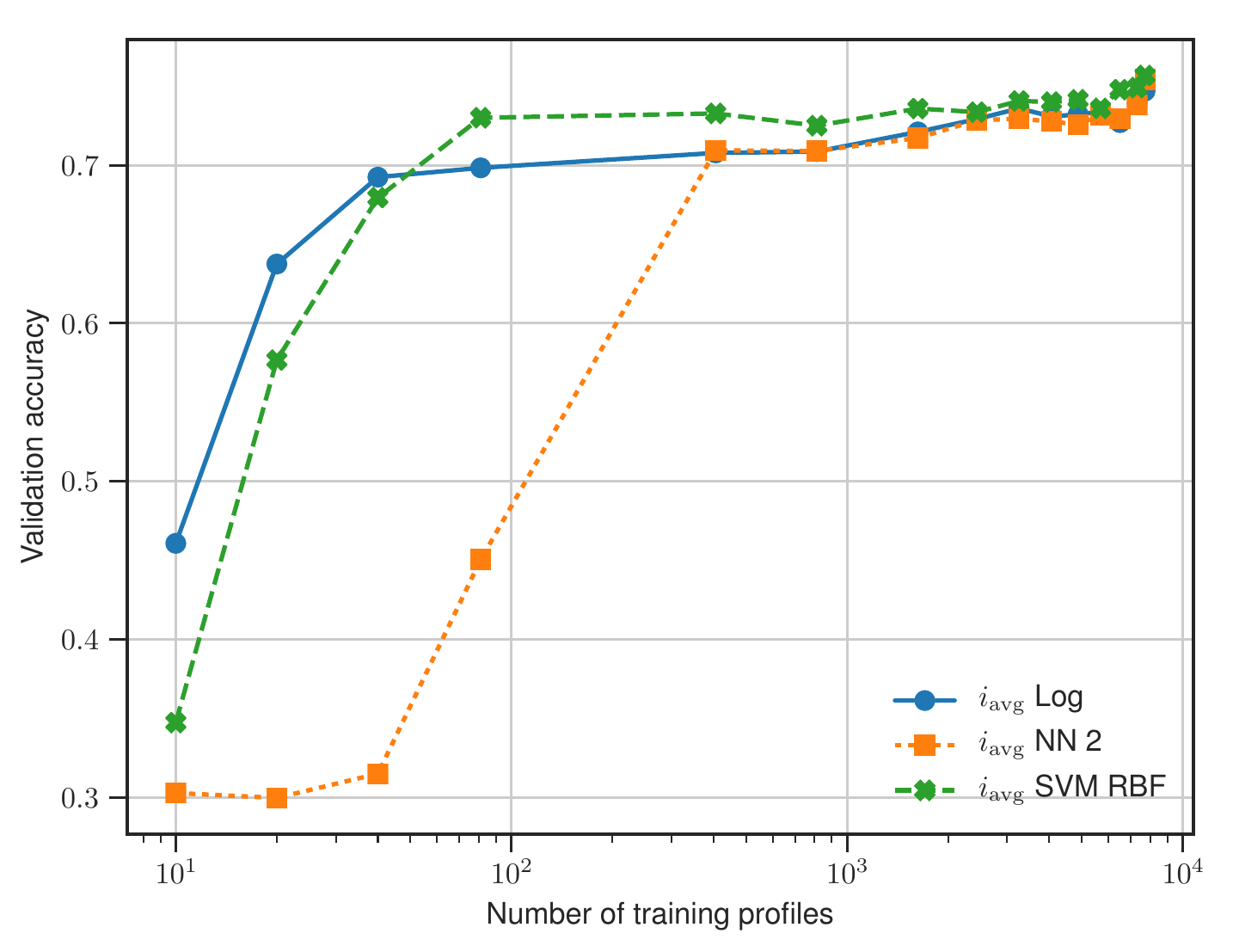}
\caption{Validation accuracy of classification models (number of correctly classified test profiles / total number of test profiles) as a function of the number of training profiles.}
\label{fig:val}
\end{figure}

Since the AUC scores were comparable, the remaining results only consider classification models fit to $\bm{i}_\text{avg}$. Models were fit using various train-to-test ratios. The train:test split was performed at random; however each model used the same random state for a given number of training profiles. The ratio of likes to dislikes was not preserved in the random splits. The training accuracy of the models is presented in Fig.~\ref{fig:train} and the validation accuracy for these models is presented in Fig.~\ref{fig:val}. The first data point represents a training size of 10 profiles and a validation size of 8,120 profiles. The last data point uses 7,723 training profiles and validation on 407 profiles (a 20:1 split). The logistic regression model (Log) and neural network (NN 2) converge to a comparable training accuracy of 0.75. Impressively, a model can have a validation accuracy greater than 0.5 after being trained on just 20 profiles. A reasonable model with a validation accuracy near 0.7 was trained on just 40 profiles.

\begin{figure}[hbt!]
\centering
\includegraphics[width=0.40\textwidth]{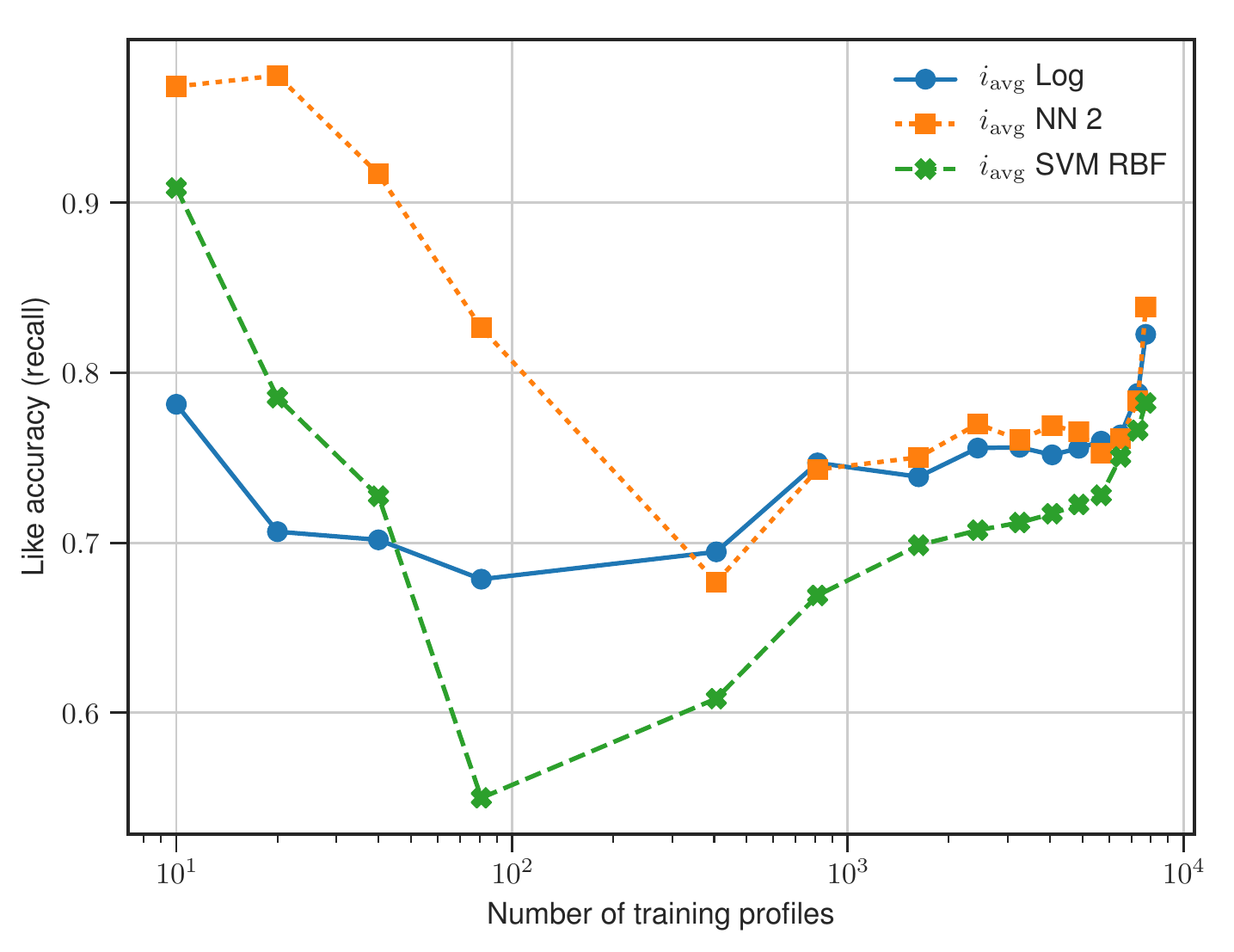}
\caption{Like accuracy (true positive rate) of classification models as a function of the number of training profiles.}
\label{fig:like}
\end{figure}

\begin{figure}[hbt!]
\centering
\includegraphics[width=0.40\textwidth]{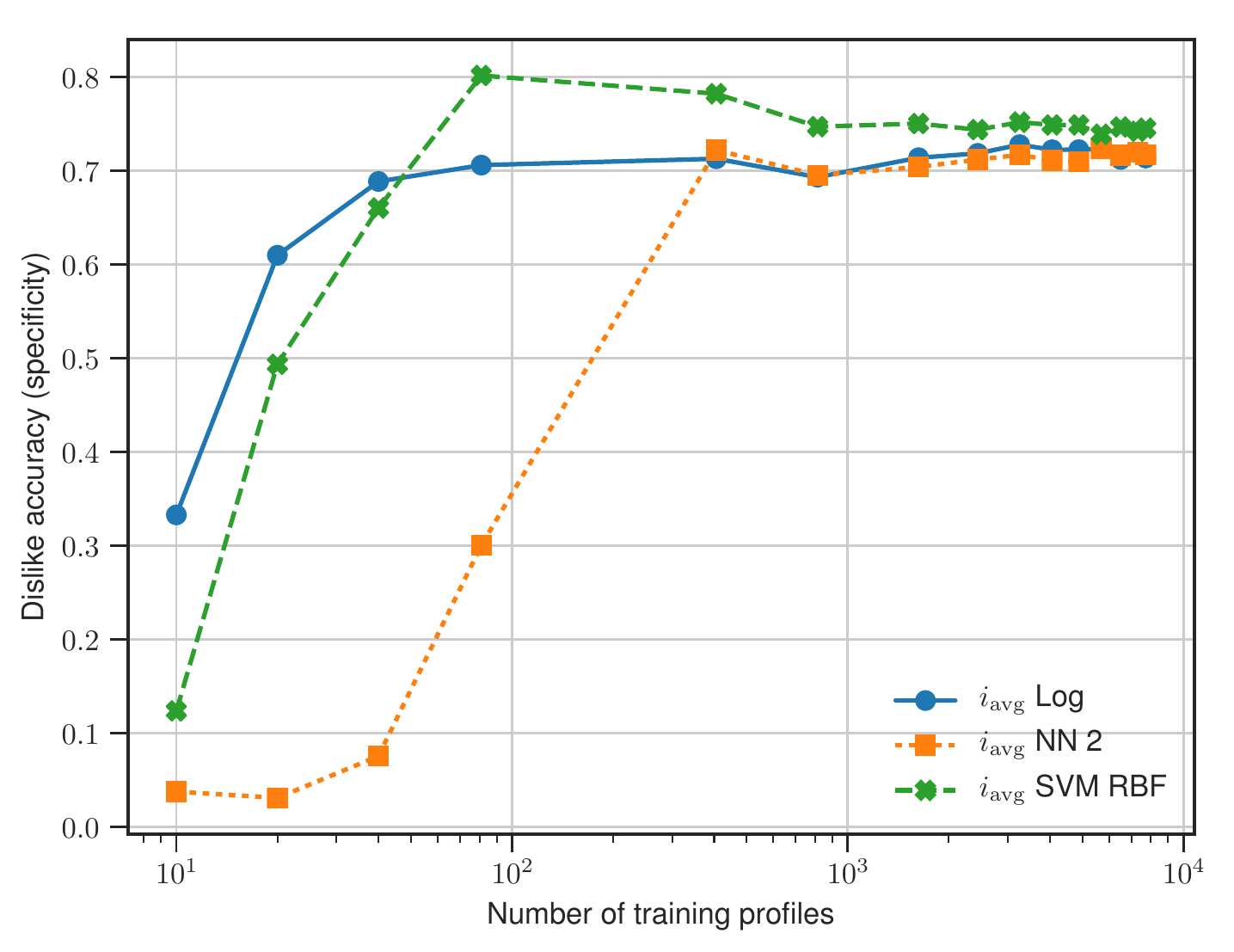}
\caption{Dislike accuracy (true negative rate) of classification models as a function of the number of training profiles.}
\label{fig:dislike}
\end{figure}

The validation accuracy can be misleading because there were significantly more disliked profiles than liked profiles. Thus it is important to consider the true positive and true negative rates to asses the model quality. The like accuracy (or true positive rate) of the models is presented in Fig.~\ref{fig:like}, and the dislike accuracy (or true negative rate) is presented in Fig.~\ref{fig:dislike}. It can be noted that the neural networks had a like bias, while the SVM has a dislike bias. The dislike bias resulted in a slightly higher validation accuracy since there were more profiles disliked than liked. Again, after 20 profiles, a reasonable classification model can be constructed. The logistic regression model trained on 20 profiles had a like accuracy of 0.7 and a dislike accuracy of 0.6. 

\begin{figure}[hbt!]
\centering
\includegraphics[width=0.45\textwidth]{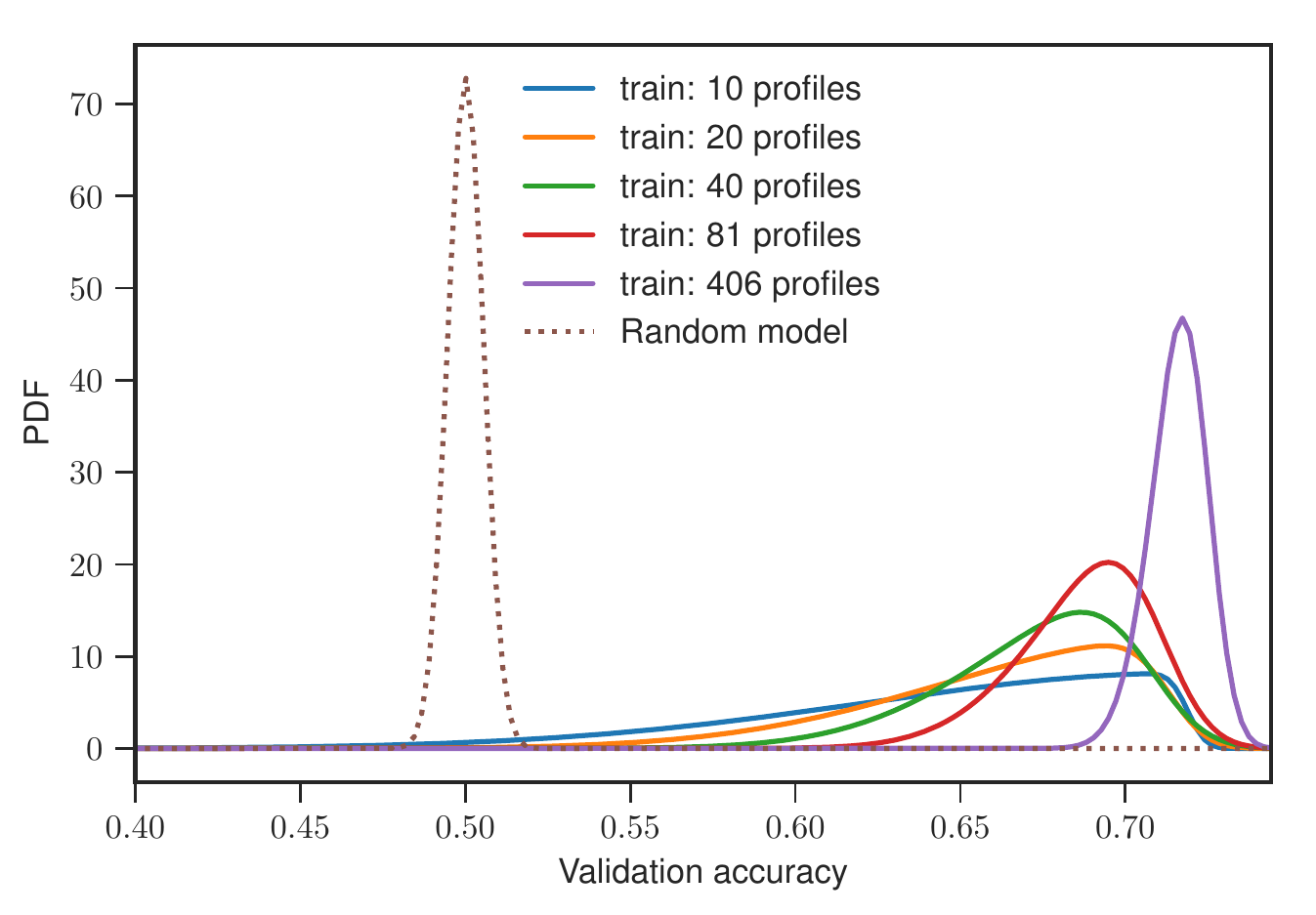}
\caption{Probability density functions (PDF) for validation accuracy of classifiers trained on either 10, 20, 40, 81, and 406 profiles. The PDF from a completely random classifier is shown as a baseline comparison.}
\label{fig:pdf}
\end{figure}

The results presented thus far could be the result of the random split. Another study is presented to better understand how the number of reviewed profiles may influence the personalized classification model. Random splits were performed 10,000 times for training sizes of 10, 20, 40, 81, and 406 profiles. The random split required at least one profile from each class (like and dislike) in order to construct a classification model. A unique logistic regression model was fit for each split and validated on the remaining test data. Again, the training of 10 profiles was validated on 8,120 profiles, and so forth for the other training sizes. The resulting validation accuracies followed a skew-normal distribution, and probability density functions (PDF) were calculated for each training size. The resulting PDFs are presented in Fig.~\ref{fig:pdf} and compared to the PDF of a completely random classifier. The validation accuracy for a completely random classifier was simulated 10,000 times and followed a normal distribution. The results demonstrate that training on just 10 dating profiles offers a significant advantage over a random classifier. The variance associated with a model's validation accuracy is shown to reduce with the number of trained profiles. This reduction of variance is significant when going from training on 81 profiles to 406 profiles.

A Python script is included in the supplementary material to calculate the the results presented here for the logistic regression model using either $\bm{i}_p$ or $\bm{i}_\text{avg}$ as the input dimension.

\section{Conclusion}

A method was presented to build personalized classification models for online dating profiles based on a user's historical preference. The method could be used to improve the user experience of online dating by reducing the time required to filter profiles. A custom data set was collected which reviewed over 8,000 Tinder profiles. Profile images containing just one face were run through a FaceNet model to extract the unique features as embeddings. Two different approaches were presented to combine these features from faces in a profile, to a unique vector representing the features of that profile. A classification model was then constructed either considering a 128 or 1280 input dimension. A simple logistic regression model was shown to find an accuracy greater than 60\% after being trained on just 20 profiles. The classification methodology continuously improves as more online dating profiles are reviewed. Additionally it was demonstrated that a classification model trained on just 10 profiles would, on average, have a much higher validation accuracy than a random classifier.

A Python command line application called \textit{tindetheus} has been released to reproduce the methodology presented in this paper. The application has three major functions: 1) Build a data set as a user browses Tinder. 2) Train a classification model to the data set. 3) Use the trained model to automatically like new Tinder profiles. Tindetheus is available online at \url{https://github.com/cjekel/tindetheus/} or \url{https://pypi.python.org/pypi/tindetheus} . 

\small{
\bibliographystyle{IEEEtranN}
\bibliography{face_and_tinder}

\begin{thebibliography}{19}
\providecommand{\natexlab}[1]{#1}
\providecommand{\url}[1]{#1}
\csname url@samestyle\endcsname
\providecommand{\newblock}{\relax}
\providecommand{\bibinfo}[2]{#2}
\providecommand{\BIBentrySTDinterwordspacing}{\spaceskip=0pt\relax}
\providecommand{\BIBentryALTinterwordstretchfactor}{4}
\providecommand{\BIBentryALTinterwordspacing}{\spaceskip=\fontdimen2\font plus
\BIBentryALTinterwordstretchfactor\fontdimen3\font minus
  \fontdimen4\font\relax}
\providecommand{\BIBforeignlanguage}[2]{{%
\expandafter\ifx\csname l@#1\endcsname\relax
\typeout{** WARNING: IEEEtranN.bst: No hyphenation pattern has been}%
\typeout{** loaded for the language `#1'. Using the pattern for}%
\typeout{** the default language instead.}%
\else
\language=\csname l@#1\endcsname
\fi
#2}}
\providecommand{\BIBdecl}{\relax}
\BIBdecl

\bibitem[Tyson et~al.(2016)Tyson, Perta, Haddadi, and Seto]{Tyson2016}
G.~Tyson, V.~C. Perta, H.~Haddadi, and M.~C. Seto, ``{A first look at user
  activity on tinder},'' in \emph{2016 IEEE/ACM International Conference on
  Advances in Social Networks Analysis and Mining (ASONAM)}, aug 2016, pp.
  461--466.

\bibitem[Abramova et~al.(2016)Abramova, Baumann, Krasnova, and
  Buxmann]{Abramova2016}
O.~Abramova, A.~Baumann, H.~Krasnova, and P.~Buxmann, ``{Gender differences in
  online dating: What do we know so far? A systematic literature review},'' in
  \emph{Proceedings of the Annual Hawaii International Conference on System
  Sciences}, vol. 2016-March, jan 2016, pp. 3858--3867.

\bibitem[Eisenthal et~al.(2006)Eisenthal, Dror, and Ruppin]{Eisenthal2006}
Y.~Eisenthal, G.~Dror, and E.~Ruppin, ``Facial attractiveness: Beauty and the
  machine,'' \emph{Neural Computation}, vol.~18, no.~1, pp. 119--142, Jan 2006.

\bibitem[Mu(2013)]{MU201359}
\BIBentryALTinterwordspacing
Y.~Mu, ``Computational facial attractiveness prediction by aesthetics-aware
  features,'' \emph{Neurocomputing}, vol.~99, no. Supplement C, pp. 59 -- 64,
  2013. [Online]. Available:
  \url{http://www.sciencedirect.com/science/article/pii/S092523121200495X}
\BIBentrySTDinterwordspacing

\bibitem[Chua et~al.(2013)Chua, Akimoto, Aguirre, and Tanaka]{Chua2013}
M.~Chua, Y.~Akimoto, H.~Aguirre, and K.~Tanaka, ``Asian female face
  classification incorporating personal attractive preference,'' in \emph{2013
  International Symposium on Intelligent Signal Processing and Communication
  Systems}, Nov 2013, pp. 413--418.

\bibitem[Gray et~al.(2010)Gray, Yu, Xu, and Gong]{Gray2010}
\BIBentryALTinterwordspacing
D.~Gray, K.~Yu, W.~Xu, and Y.~Gong, \emph{Predicting Facial Beauty without
  Landmarks}.\hskip 1em plus 0.5em minus 0.4em\relax Berlin, Heidelberg:
  Springer Berlin Heidelberg, 2010, pp. 434--447. [Online]. Available:
  \url{https://doi.org/10.1007/978-3-642-15567-3_32}
\BIBentrySTDinterwordspacing

\bibitem[Rothe et~al.(2016)Rothe, Timofte, and Gool]{Rothe2016}
R.~Rothe, R.~Timofte, and L.~V. Gool, ``Some like it hot!; visual guidance for
  preference prediction,'' in \emph{2016 IEEE Conference on Computer Vision and
  Pattern Recognition (CVPR)}, June 2016, pp. 5553--5561.

\bibitem[Zang et~al.(2017)Zang, Yamasaki, Aizawa, Nakamoto, Kuwabara, Egami,
  and Fuchida]{Zang2017}
X.~Zang, T.~Yamasaki, K.~Aizawa, T.~Nakamoto, E.~Kuwabara, S.~Egami, and
  Y.~Fuchida, ``{Prediction of users' facial attractiveness on an online dating
  website},'' in \emph{2017 IEEE International Conference on Multimedia Expo
  Workshops (ICMEW)}, jul 2017, pp. 255--260.

\bibitem[Parkhi et~al.(2015)Parkhi, Vedaldi, and Zisserman]{Parkhi15}
O.~M. Parkhi, A.~Vedaldi, and A.~Zisserman, ``{Deep Face Recognition},'' in
  \emph{British Machine Vision Conference}, 2015.

\bibitem[Huang et~al.(2007)Huang, Ramesh, Berg, and Learned-Miller]{LFWTech}
G.~B. Huang, M.~Ramesh, T.~Berg, and E.~Learned-Miller, ``{Labeled faces in the
  wild: A database for studying face recognition in unconstrained
  environments},'' University of Massachusetts, Amherst, Tech. Rep. 07-49, oct
  2007.

\bibitem[Schroff et~al.(2015)Schroff, Kalenichenko, and
  Philbin]{Schroff_2015_CVPR}
F.~Schroff, D.~Kalenichenko, and J.~Philbin, ``{FaceNet: A Unified Embedding
  for Face Recognition and Clustering},'' in \emph{The IEEE Conference on
  Computer Vision and Pattern Recognition (CVPR)}, jun 2015.

\bibitem[Wen et~al.(2016)Wen, Zhang, Li, and Qiao]{Wen2016}
\BIBentryALTinterwordspacing
Y.~Wen, K.~Zhang, Z.~Li, and Y.~Qiao, ``{A Discriminative Feature Learning
  Approach for Deep Face Recognition},'' in \emph{Computer Vision -- ECCV 2016:
  14th European Conference, Amsterdam, The Netherlands, October 11--14, 2016,
  Proceedings, Part VII}, B.~Leibe, J.~Matas, N.~Sebe, and M.~Welling,
  Eds.\hskip 1em plus 0.5em minus 0.4em\relax Cham: Springer International
  Publishing, 2016, pp. 499--515. [Online]. Available:
  \url{https://doi.org/10.1007/978-3-319-46478-7_31}
\BIBentrySTDinterwordspacing

\bibitem[Amos et~al.(2016)Amos, Ludwiczuk, and
  Satyanarayanan]{amos2016openface}
B.~Amos, B.~Ludwiczuk, and M.~Satyanarayanan, ``{OpenFace: A general-purpose
  face recognition library with mobile applications},'' CMU-CS-16-118, CMU
  School of Computer Science, Tech. Rep., 2016.

\bibitem[Guo et~al.(2016)Guo, Zhang, Hu, He, and Gao]{guo2016}
Y.~Guo, L.~Zhang, Y.~Hu, X.~He, and J.~Gao, ``M{S}-{C}eleb-1{M}: A dataset and
  benchmark for large scale face recognition,'' in \emph{European Conference on
  Computer Vision}.\hskip 1em plus 0.5em minus 0.4em\relax Springer, 2016.

\bibitem[Szegedy et~al.(2016)Szegedy, Ioffe, Vanhoucke, and Alemi]{Szegedy2016}
\BIBentryALTinterwordspacing
C.~Szegedy, S.~Ioffe, V.~Vanhoucke, and A.~Alemi, ``Inception-v4,
  inception-resnet and the impact of residual connections on learning,'' in
  \emph{Proceedings of the Thirty-First {AAAI} Conference on Artificial
  Intelligence, February 4-9, 2017, San Francisco, California, {USA.}}, S.~P.
  Singh and S.~Markovitch, Eds.\hskip 1em plus 0.5em minus 0.4em\relax {AAAI}
  Press, 2016, pp. 4278--4284. [Online]. Available:
  \url{http://arxiv.org/abs/1602.07261}
\BIBentrySTDinterwordspacing

\bibitem[Zhang et~al.(2016)Zhang, Zhang, Li, and Qiao]{Zhang2016}
K.~Zhang, Z.~Zhang, Z.~Li, and Y.~Qiao, ``{Joint Face Detection and Alignment
  Using Multitask Cascaded Convolutional Networks},'' \emph{IEEE Signal
  Processing Letters}, vol.~23, no.~10, pp. 1499--1503, oct 2016.

\bibitem[Pedregosa et~al.(2011)Pedregosa, Varoquaux, Gramfort, Michel, Thirion,
  Grisel, Blondel, Prettenhofer, Weiss, Dubourg, Vanderplas, Passos,
  Cournapeau, Brucher, Perrot, and Duchesnay]{scikit-learn}
F.~Pedregosa, G.~Varoquaux, A.~Gramfort, V.~Michel, B.~Thirion, O.~Grisel,
  M.~Blondel, P.~Prettenhofer, R.~Weiss, V.~Dubourg, J.~Vanderplas, A.~Passos,
  D.~Cournapeau, M.~Brucher, M.~Perrot, and E.~Duchesnay, ``{Scikit-learn:
  Machine Learning in {P}ython},'' \emph{Journal of Machine Learning Research},
  vol.~12, pp. 2825--2830, 2011.

\bibitem[Chollet and Others(2015)]{chollet2015keras}
F.~Chollet and Others, ``{Keras},'' \url{https://github.com/fchollet/keras},
  2015.

\bibitem[Abadi and {et al.}(2015)]{tensorflow2015-whitepaper}
\BIBentryALTinterwordspacing
M.~Abadi and {et al.}, ``{TensorFlow}: Large-scale machine learning on
  heterogeneous systems,'' 2015, software available from tensorflow.org.
  [Online]. Available: \url{https://www.tensorflow.org/}
\BIBentrySTDinterwordspacing

\end{thebibliography}
}
\end{document}